\documentclass[twoside,11pt]{article} 
\usepackage{jmlr2e}
\usepackage{todonotes}
\usepackage{float}
\usepackage{amssymb}	
\usepackage{graphicx}  
\usepackage{algorithm} 
\usepackage[noend]{algpseudocode} 
\usepackage{hyperref}
\hypersetup{
    colorlinks=true,
    linkcolor=blue,
    filecolor=magenta,      
    urlcolor=cyan
}

\usepackage{url}
\usepackage{subfigure}
\usepackage{tabularx}
\usepackage{xfrac}

\newcommand{\be}{\begin{equation}}
\newcommand{\ee}{\end{equation}}

\newcommand{\mb}{\mathbf}

\usepackage{amsmath} \DeclareMathOperator*{\argmax}{argmax}

\renewcommand{\Comment}[2][.65\linewidth]{%
  \leavevmode\hfill\makebox[#1][l]{$\triangleright$~#2}}
\algnewcommand\algorithmicto{\textbf{to}}
\algnewcommand\RETURN{\State \textbf{return} }

\begin{document}

\title{Real-time Approximate Bayesian Computation for Scene Understanding}

\author{\name Javier Felip \email javier.felip.leon@intel.com
        \AND  
        \name Nilesh Ahuja \email nilesh.ahuja@intel.com
        \AND  
        \name David G\'omez-Guti\'errez \email david.gomez.gutierrez@intel.com
        \AND  
        \name Omesh Tickoo \email omesh.tickoo@intel.com \\
        \addr Intel \\
        \name Vikash Mansinghka \email vkm@mit.com \\
        \addr MIT
         }
         
\editor{ }

\maketitle

\begin{abstract}
Consider scene understanding problems such as predicting
where a person is probably reaching, or inferring the pose of
3D objects from depth images, or inferring the probable street
crossings of pedestrians at a busy intersection. This paper shows
how to solve these problems using Approximate Bayesian Computation. The underlying generative models are built from realistic
simulation software, wrapped in a Bayesian error model for the gap
between simulation outputs and real data. The simulators are drawn
from off-the-shelf computer graphics, video game, and traffic
simulation code. The paper introduces two techniques for speeding up
inference that can be used separately or in combination. The first is to train neural surrogates of the simulators,
using a simple form of domain randomization to make the surrogates more robust to the gap between the simulation and reality. The second is to adaptively discretize the latent variables using a Tree-pyramid approach adapted from computer graphics. This paper 
also shows performance and accuracy measurements on real-world problems,
establishing that it is feasible to solve these problems in real-time.
\end{abstract}

\section{Introduction}

Visual scene understanding involves estimating quantities of interest (such as object category, location, etc.) from observed images and videos. Bayesian approaches to such problems are gaining importance as these provide uncertainty estimates of the variables being estimated. However, these tend to be expensive and unfeasible for real-time implementation. This paper presents a framework for efficiently solving such problems in real-time using Approximate Bayesian Computation (ABC)~\citep{toni2009approximate, sunnaaker2013approximate, beaumont2002approximate, marin2012approximate}. The underlying generative models needed for ABC are built from realistic off-the-shelf simulation software but additionally include a simple Bayesian error model for the gap between simulation outputs and real data.  

\begin{figure*}
	\centering
	\includegraphics[width=0.99\textwidth]{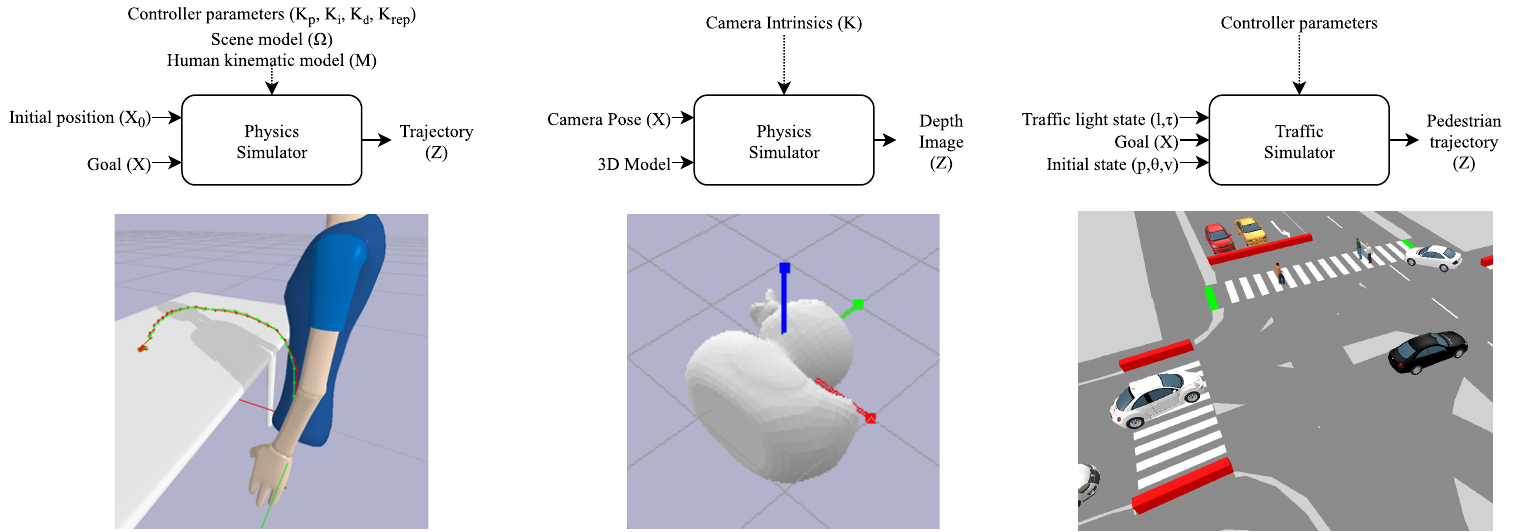}
	\caption{{\bf Simulators used to define generative models:} This paper shows results for reach-intent estimation (left), 3D object pose estimation (center), and pedestrian intent estimation (right). The top row shows the inputs, outputs, and parameters (assumed known) for each simulator. The bottom row shows the visualization of a sample output from each simulator. The inference task is to infer probable inputs given real-world data that is treated as a noise-corrupted simulator output.}	
	\label{fig:sim_schematics}
\end{figure*}

ABC itself can be computationally challenging~\citep{wegmann2009efficient}: Monte Carlo approximations require expensive likelihood evaluations, and the posterior distributions can be complex and multimodal. This paper introduces two techniques, that together make it possible to apply ABC with complex simulators in real-time applications. The first technique is to train neural surrogates for the simulators. The neural surrogates allow for fast approximate likelihood evaluations, which can be many orders of magnitude faster than the original simulators. Unlike the neural surrogates that are more frequently used in ABC, these surrogates are trained on simulated data that is intentionally corrupted with additive Gaussian noise, following the Bayesian error model. This helps to account for the gap between the simulator and reality. It can be viewed as a simple form of domain randomization~\citep{tobin2017domain}, an increasingly popular technique for building deep learning systems that are trained on synthetic data from complex simulators but deployed on data from the real world~\citep{li2018vision}. However, even with neural surrogates, our experiments confirm that Monte Carlo approximations to these kinds of complex posteriors can still be unnecessarily expensive. Thus, the second technique is an adaptive discretization algorithm for approximating the posterior distribution based on Tree-Pyramids (TPs), a hierarchical data structure from computer graphics~\citep{fuchs1980visible}. Experiments show that this TP-based approximate inference algorithm can represent complex posterior distributions induced by neural surrogates more accurately than Monte Carlo approximations, given the same computational budget.

The approach taken in this paper is complementary to recent work on
Gaussian process surrogates~\citep{liu2014gaussian, papamakarios2016fast} and synthetic neural likelihoods
for ABC~\citep{papamakarios2018sequential}. These papers focus on
reducing the number of simulations needed for posterior parameter
estimation in queuing models and differential equation models of
biological systems. By contrast, our approach introduces techniques
suitable for real-time applications in scene understanding. For such applications, there is no need to minimize the number of simulations used to train surrogates, as one surrogate will be reused to interpret data from many scenarios. We are thus able to use a simple approach to train neural surrogates. 

To the best of our knowledge, this is the first paper that combines ABC, deep learning, domain randomization, and hierarchical sampling using Tree-Pyramids (TPs). It is also the first to show that ABC approaches can be practical for real-time scene understanding. Indeed, existing ``analysis by synthesis'' approaches to vision problems, for instance~\cite{yuille2006vision, mansinghka2013approximate, kulkarni2015picture}, were offline architectures, too slow for real-time applications. 

The outline of the paper is as follows: in Section \ref{sec:Framework}, an overview of the proposed ABC framework is presented and techniques employed to achieve real-time performance are described. Section \ref{sec:Applications} shows the application of this framework to three diverse scene-understanding problems along with empirical results for each. Finally, conclusions are presented in Section \ref{sec:Conclusions}.

\section{Real-time Approximate Bayesian Computations}
\label{sec:Framework}

The real-time scene understanding results in this paper are based on ABC in probabilistic generative models defined from simulation software, e.g. see Figure~\ref{fig:sim_schematics}. The generative process that induces these models fits the usual template for ABC, but with a Bayesian model for the gap between simulation outputs and real data. The model variables and the accompanying notation is introduced next.

\subsection{Generative modeling via realistic simulators}
It is desired to estimate a latent variable, $X$, from an observation, $D$. Assume that we have access to a simulator that models the transformation from the input space of $X$ to the observation space. The distribution on $X$ is represented by a scene prior, $p(X)$. Given an input $X$, the simulator produces a simulated output, $Z$, which has an associated conditional distribution $p(Z|X)$. The relation between $D$ and $Z$ is modeled as:
\begin{equation}
\begin{aligned}
p(D|Z, \epsilon)  &\sim  Normal(Z, \epsilon), \\
p(\epsilon) &\sim Gamma(1,1),
\end{aligned}
\end{equation}
where, $\epsilon$ is a slack variable representing the gap between simulation and reality. Scene understanding then becomes the problem of performing posterior inference jointly over $\epsilon$ and $X$:
\be
\label{eq:posterior}
p(X,\epsilon | D) \propto p(X) p(\epsilon) p(D|X, \epsilon).
\ee
As is typical in ABC problems, the likelihood distribution induced by the simulators in this paper, $p(D|X, \epsilon) = \int dZ p(D|Z, \epsilon) P(Z|X)$, can be difficult to calculate.

Note that we are interested in posterior inferences about both the scene $X$ and the parameter $\epsilon$ that governs the error. This is because we may wish to filter out scenes where 
\be
\label{eq:slack}
\epsilon^* = \argmax p(\epsilon | D)
\ee
is high. The dynamics of joint real-time inference over $X$ and $\epsilon$ implement a form of real-time model criticism~\citep{ratmann2009model}. When the data is unlikely under the simulator, high values of $\epsilon$ are favored in the posterior. In addition to improving robustness to simulator-data mismatch, this extension makes it possible to filter datasets to detect samples that violate the assumptions in a given simulator model. In Section \ref{sec:Applications}, we show how this kind of filtering can find anomalies such as pedestrians who break traffic rules. The following two subsections describe the techniques used to achieve real-time performance.

\subsection{Training neural surrogates with domain randomization}

\begin{figure}
	\centering
	\includegraphics[width=0.5\textwidth]{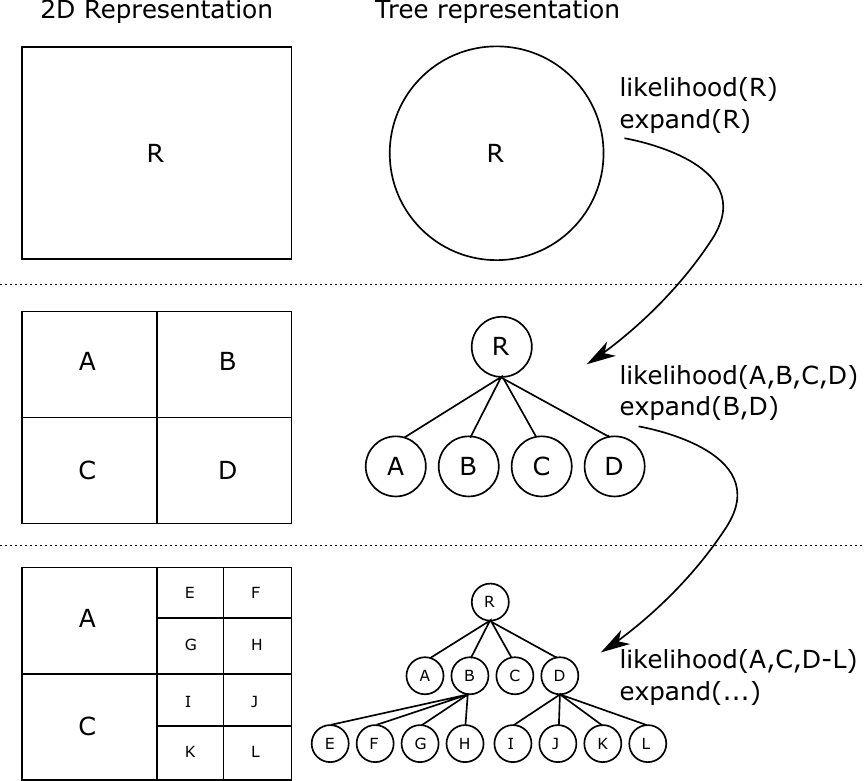}
	\caption{{\bf Adaptive discretization of 2D latent variables using a 2D-TP, namely a quadtree.} This figure shows an example of the first subdivisions of the sampling algorithm for the 2D case, the construction of the TP, and the geometry of each node in relation to the 2D space being sampled.}
	\label{fig:quadtree_example}
\end{figure}

The first technique used in this paper to make real-time ABC feasible is to train neural surrogates $f_\phi(X)$ for the simulator outputs. These are deterministic neural networks that only approximate the true underlying simulator $p(Z|X)$ using their parameters $\phi$:

$$f_\phi(X) \approx \argmax_Z p(Z|X).$$

However, unlike many applications of ABC in science and engineering, our scene understanding involve real-world data that may not be accurately modeled by the underlying simulator. We thus need a training approach that is robust to the mismatch between the simulator and the true data generating process. We employ a model-based variation on domain randomization, a recently introduced technique that is in widespread use for applying deep learning to synthetic data~\citep{tobin2017domain}. Specifically, we train a randomized surrogate $f_\phi^*$ to replicate the most probable outcome of a noisy version of the simulator that includes the error model, with a high value $\epsilon^*$ of Gaussian error added to the simulator outputs: 
$$
f_\phi^*(X) \approx \argmax_D p(D|X, \epsilon^*).
$$
In multiple experiments, this randomized training approach yielded surrogates that performed more robustly on real-world data. This kind of simple, noise-based domain randomization was also found to improve robustness in other applications involving physical simulators~\citep{li2018vision}. We note that there are intriguing parallels between this domain randomization and the error model; future work exploring this relationship may be relevant.

Given this neural surrogate, the inference problem we solve is defined by the following approximation to the true posterior (following eq. (\ref{eq:posterior})):
\be
p(X,\epsilon|D) \propto p(X)p(\epsilon)Normal(f_\phi^*(X),\epsilon).
\label{eq:surrogate_posterior}
\ee

Training data for the surrogates are generated by their respective simulators, i.e. the input to the network is $X$, and the output of the network is $Z$ generated by the simulator. The networks are trained by mini-batch stochastic gradient-descent (SGD) and hyper-parameter optimization to minimize the following loss function on synthetic data elements $(\hat{X}, \hat{Z})$ sampled from the generative model:
\be
L=\|f_{\phi}(\hat{X})-Normal(\hat{Z}|\hat{X},\epsilon^*)\|^2.
\ee

\begin{algorithm*}
	\caption{Inference algorithm based on adaptive discretization using posterior TP}
	\label{alg:TP}
	\begin{algorithmic}[1]
		\Function{computeTPPosteriorApproximation}{$k, \tau, \rho, D, c, r$}
		\State $T \leftarrow TP(k, c, r)$ \Comment{Initialize tree with the dimensions, center and radius.}
		\State $\chi \leftarrow \{getRoot(T)\}$	\Comment{Add the tree root node to the expansion set $\chi$.}
		\While{$\chi \neq \emptyset$} 
		\State $\{ e_n \} \leftarrow genCandidateTPExpansions(T, \chi, k)$ 
		\State $\chi \leftarrow \emptyset$ 
		\State $L \leftarrow likelihood(e_i, D)$ \Comment{Compute likelihoods for each candidate node.}
		\For{$n \leftarrow 1...N_L$}
		\If{$L_n > \tau$ {\bf and} $getRadius(e_n) > \rho$}
		\State $\chi = \chi \cup \{e_n\}$ \Comment{Add nodes from the candidate set E that satisfy}
		\State \Comment{the expansion criteria  to the expansion set $\chi$.}
		\EndIf
		\EndFor
		\EndWhile
		\State \Return $T, \mu$	
		\EndFunction
	\end{algorithmic}
\end{algorithm*}

\label{sec:TP}
\begin{algorithm*}
	\caption{Leaf expansion algorithm for a k-Dimensional Tree Pyramid}\label{alg:TP_expand}
	\label{alg:expansion}
	\begin{algorithmic}[1]
		\Function{genCandidateTPExpansions}{$T,\chi,k$}
		\State $E \leftarrow \emptyset$
		\For{$n \leftarrow 1...N_\chi$}
		\State $c \leftarrow getCenter(\chi_n)$ \Comment{Obtain center and radius of the node to expand.}
		\State $r \leftarrow getRadius(\chi_n)$
		\State $\delta_c \leftarrow product([1,-1], k) \cdot \frac{r}{2}$ \hfill\makebox[0.6\linewidth][l]{$\triangleright$Cartesian product of +r and -r in $2^k$-dimensions.}
		\State $new_c \leftarrow c + \delta_c$ \hfill\makebox[0.6\linewidth][l]{$\triangleright$ Obtain the centers of the new $2^k$ nodes.}
		\State $new_r \leftarrow r/2$
		\State $E = E \cup \{new_c, new_r\}$
		\EndFor
		\State $T \leftarrow insert(T,E)$
		\State \Return $N$
		\EndFunction
	\end{algorithmic}
\end{algorithm*}

\subsection{Adaptive discretization via Tree Pyramids}
A second technique we use to speed up inference is to adaptively discretize the latent variables, using techniques from computer graphics. Specifically, we propose a k-dimensional tree-pyramid (KD-TP) based posterior sampling algorithm. A KD-TP is a full tree where each node has either zero or $2^K$ children and represents a hierarchical subdivision of k-dimensional space into convex subspaces; in this paper, as imposed by the TP structure, the divisions of the space are always by half on each dimension. By hierarchically sampling the posterior in each of the subspaces, we show an efficient sampling approach. See an example of the creation of the tree and the sampling procedure in Figure~\ref{fig:quadtree_example}.

Algorithm~\ref{alg:TP} shows our adaptive discretization strategy for posterior sampling using a Tree Pyramid. The required parameters consist of the following: (i) $k$, dimensionality of the space; (ii) $\tau$, the likelihood expansion threshold; (iii) $\rho$, the resolution limit; (iv) $D$, the observation; (v) $\epsilon$, the slack term; (vi) $c$, the sample space center/root node location; and (vii) $r$, the sample space radius/ root node span.

The procedure starts by adding the tree root node to the expansion set $\chi$. While there are nodes in the expansion set, the nodes are expanded (using Algorithm~\ref{alg:expansion}) and their children are added to the evaluation set $E$. Using the observed data $D$ and the discretized slack terms $\Sigma$ the likelihood for each node in $E$ is computed (sample values are obtained as the center position of each node). It is important to note that the use of neural emulators enables batch computations of likelihoods. All nodes $n \in E$ with each $\epsilon \in \Sigma$ are processed in a single call dramatically reducing the computation time.
In lines 8 to 11, the samples from $E$ with a radius bigger than the limit radius $\rho$ and with a likelihood over the threshold $\tau$ are added to the set $\chi$. The likelihood of each sample is the maximum of the likelihoods obtained with each of the slack values $\epsilon \in \Sigma$. When there are no more nodes to be expanded, the result of the inference is the center of the leaf node with maximum likelihood.

\begin{figure*}[h]
	\centering
	\includegraphics[trim={14cm 12cm 30cm 1cm}, clip, width=0.23\textwidth]{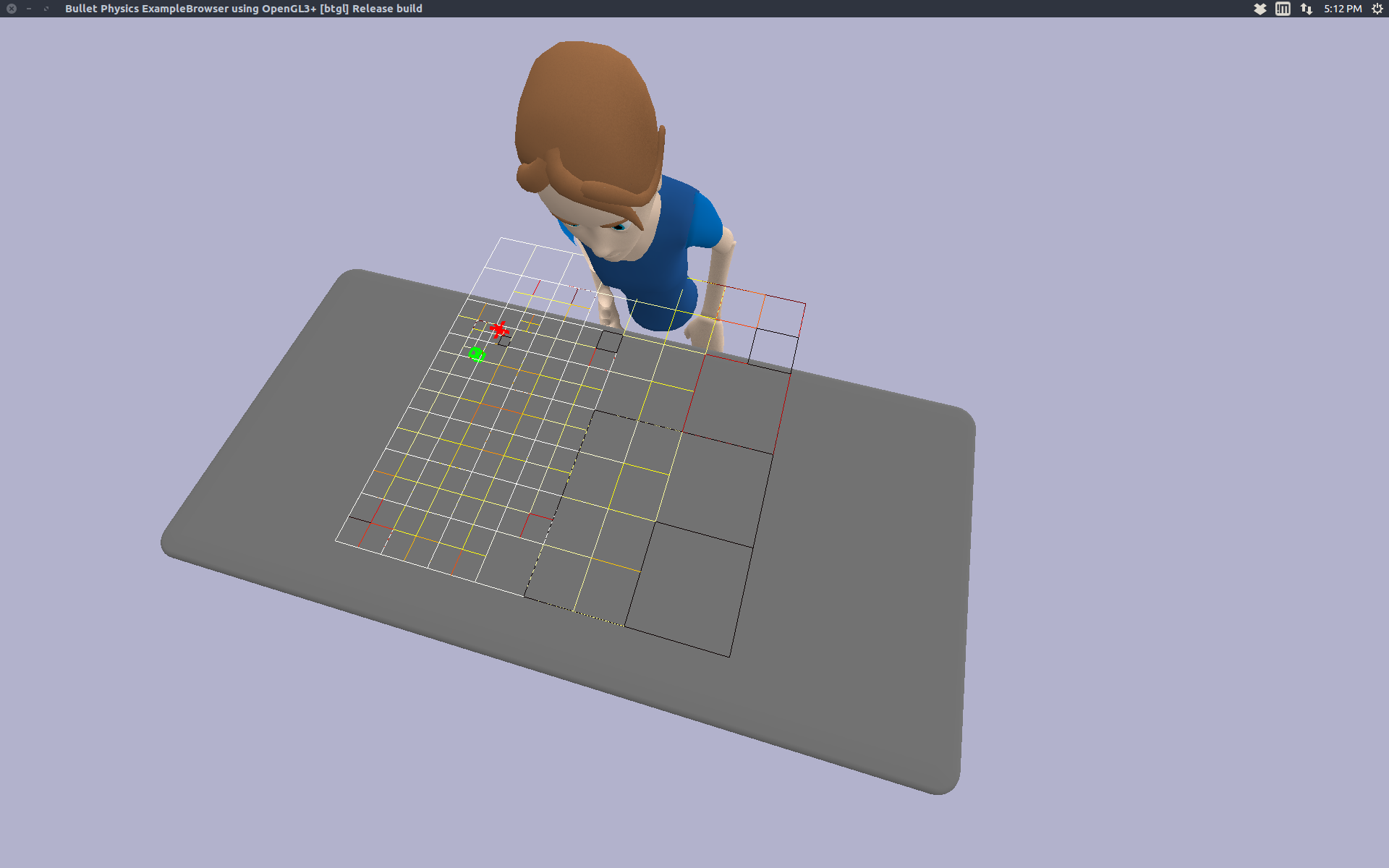}
	\includegraphics[trim={14cm 12cm 30cm 1cm}, clip, width=0.23\textwidth]{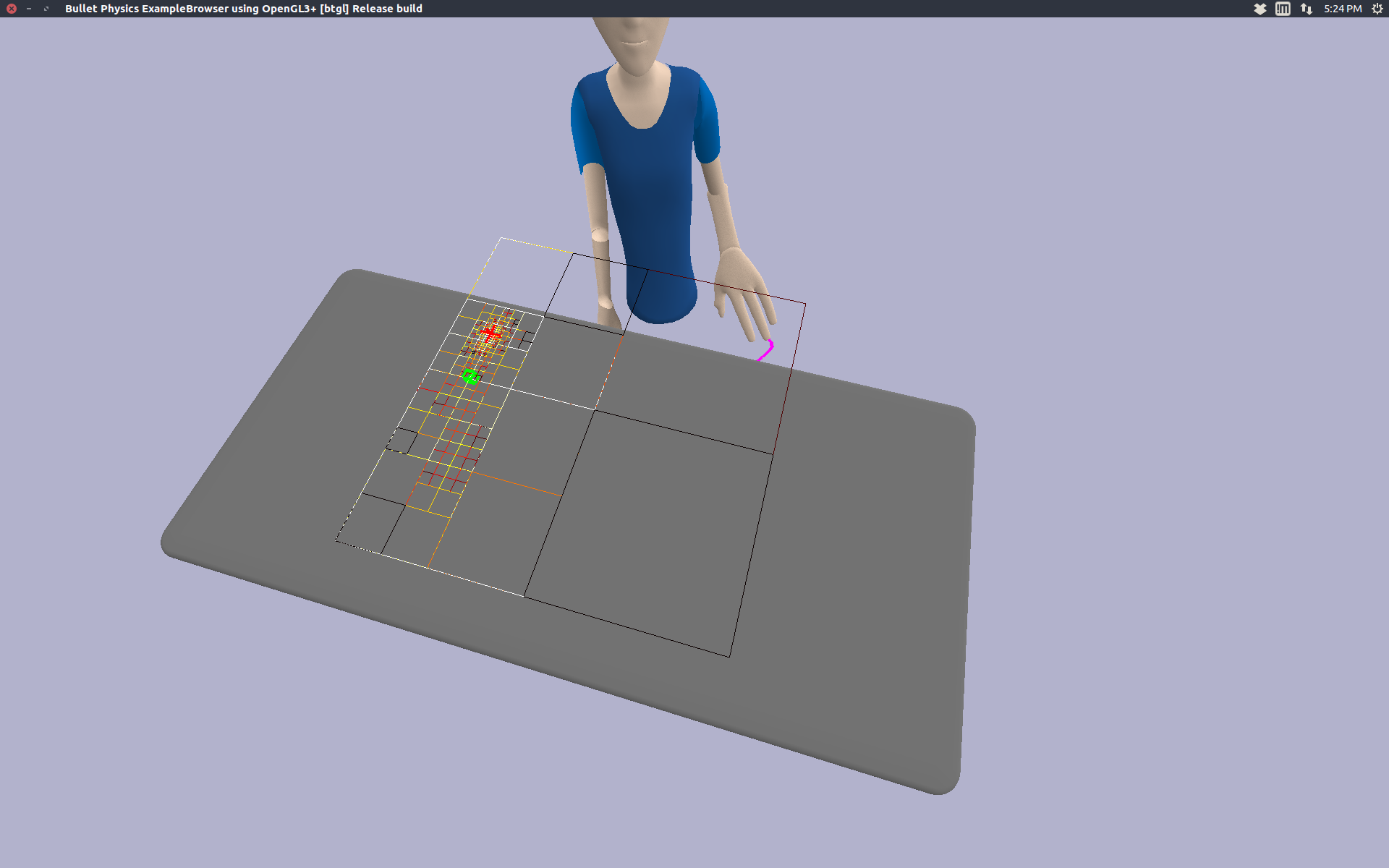}
	\includegraphics[trim={14cm 12cm 30cm 1cm}, clip, width=0.23\textwidth]{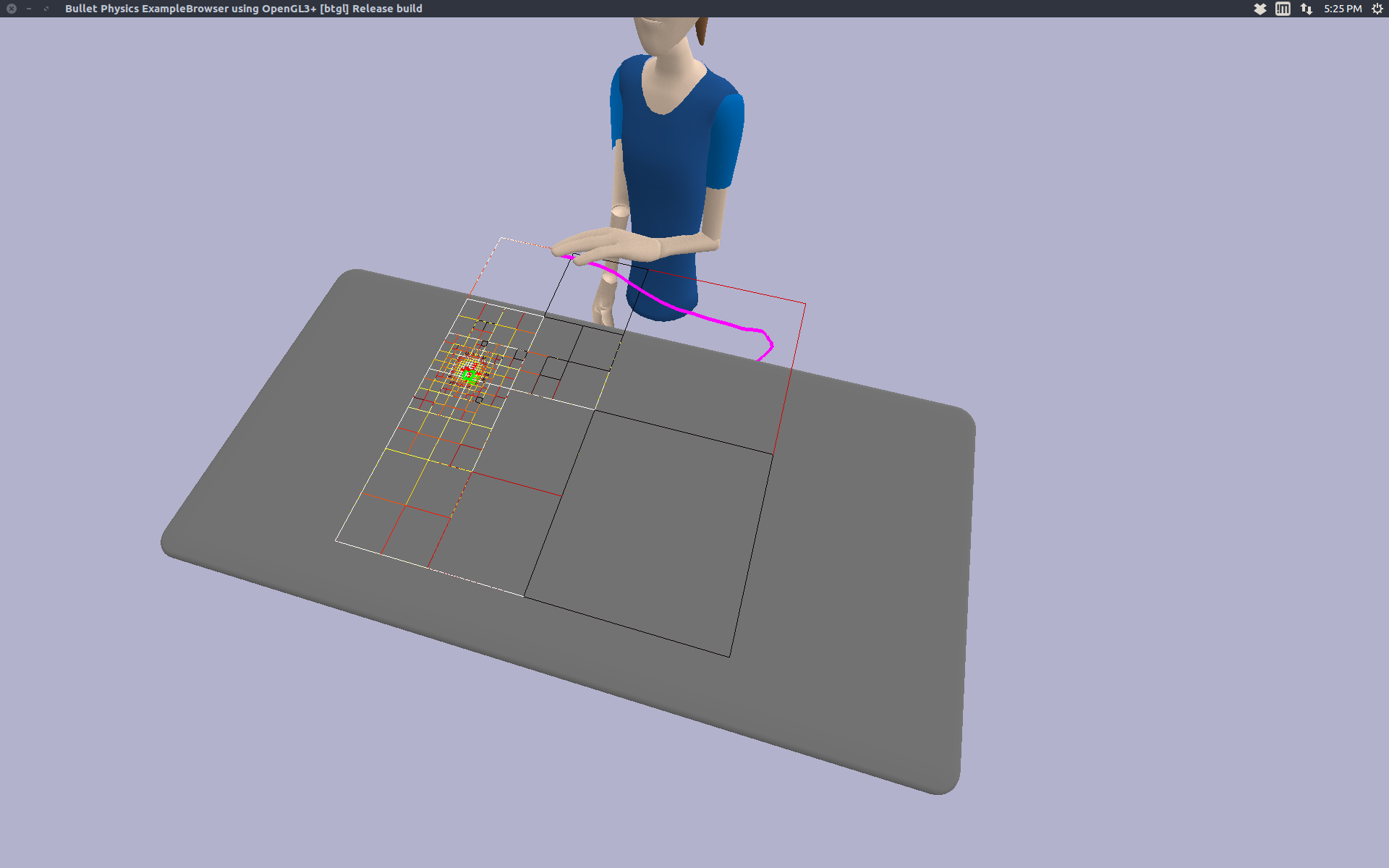}
	\includegraphics[trim={6.15cm 3cm 10cm 0.4cm}, clip, width=0.23\textwidth]{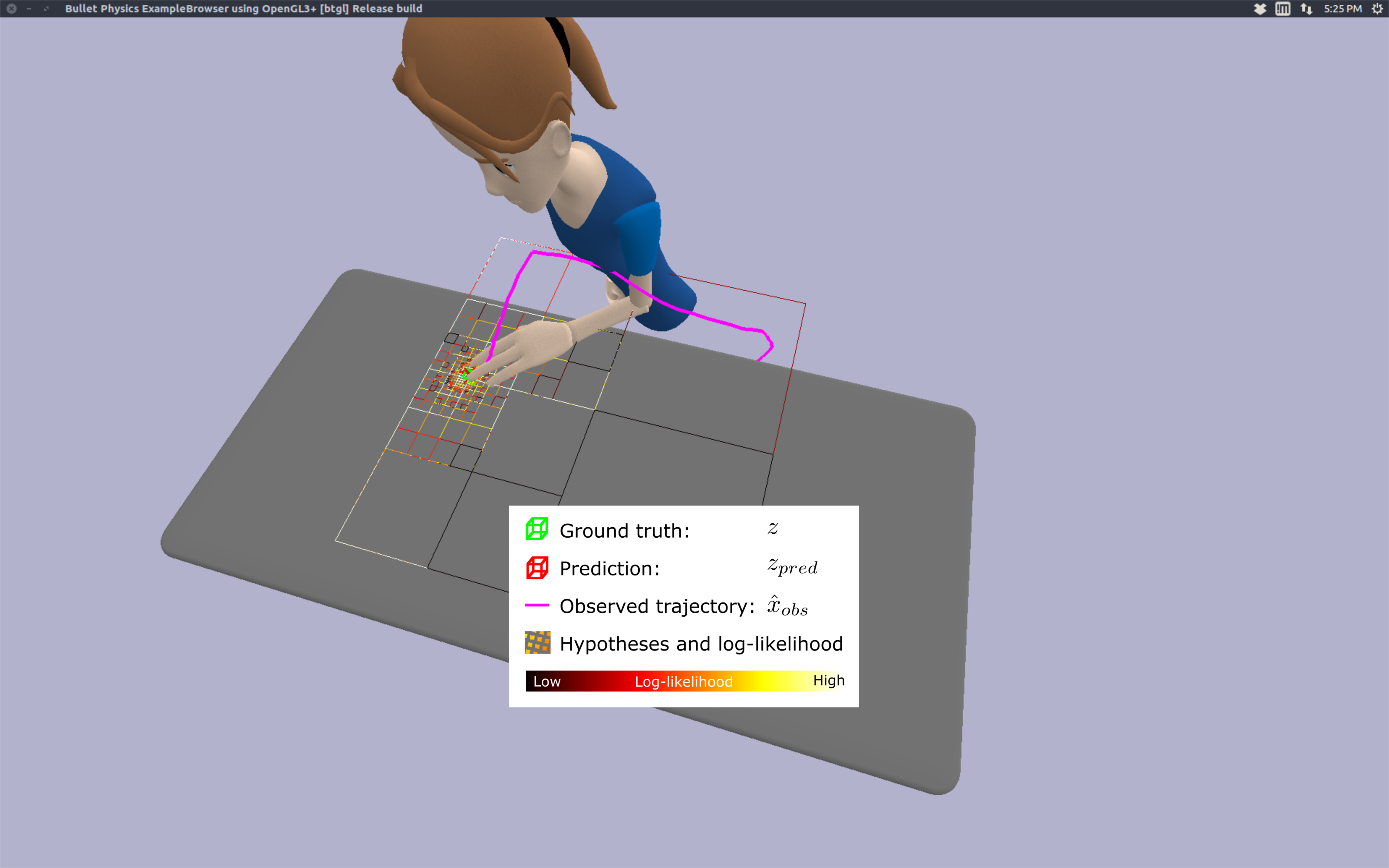}
	\caption{{\bf Adaptive discretization of latent variables using Tree-Pyramids (TPs).} This figure shows an approximate posterior distribution for the reaching intent application at four different moments in a single reaching movement. Note that the tree grows articulated in regions of high probability.}
	\label{fig:quadtree}
	\vspace{-20pt}
\end{figure*}

Figure~\ref{fig:quadtree} shows an example of executing the proposed posterior sampling algorithm in a 2D-TP, more commonly referred to as a quadtree.


\section{Application to Scene-Understanding problems}
\label{sec:Applications}
\begin{figure*}[htbp]
	\centering
	\includegraphics[height=.8in]{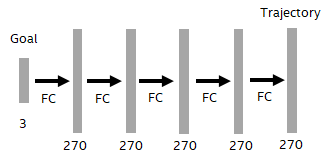} \hspace{4cm}
	\includegraphics[height=.8in]{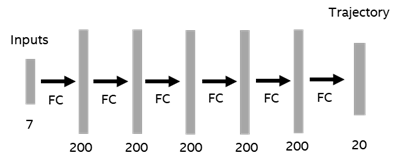}
	\includegraphics[height=.8in]{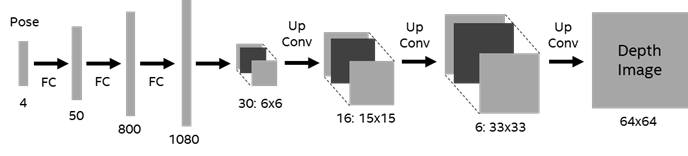}
	\caption{{Neural surrogate architectures used for real-time approximate likelihood evaluation.} Architectures are shown for reaching intent estimation (top left), pedestrian goal estimation in traffic (top right), and 3D pose estimation (bottom).}
	\label{fig:neural_archs}
	\vspace{-10pt}
\end{figure*}

To demonstrate the effectiveness of the proposed method, we focus on three scene understanding applications: (a) predicting what physical object a person is reaching for, (b) inferring the probable street crossing of pedestrians at a traffic intersection, and (c) estimating pose of 3D objects from depth images. The architectures of the neural surrogates for the three target applications are shown in Figure~\ref{fig:neural_archs}. Empirical results for each of these applications are presented in this section.

\subsection{Reaching intent model}
\paragraph{Generative Model}
For reach intent prediction, the latent variable to be estimated $X \in \mathbb{R}^3$ is the reaching goal coordinate, constrained to a table plane. The observation $D \in \mathbb{R}^{90 \times 3}$ is 90 samples of the hand trajectory at 30Hz.

The generative model consists of a kinematic model of the task environment, and a 9 Degrees Of Freedom (DOF) model of a human torso and left arm whose joint limits and dimensions that of an average US adult\footnote{Source: National Center for Health Statistics. Anthropometric Reference Data for Children and Adults: United States, 2011-2014}. 
The kinematic model is loaded into a physics engine~\citep{bullet} and trajectories are generated by simulating 3 seconds of hand trajectory at 30Hz, for trajectories that reach the goal in less than 3 seconds, the hand mantains the goal position for the remaining sampling time.
The arm motion is obtained by a Jacobian pseudo-inverse $J^{\dagger}$ that transforms task space velocities $\dot{x}$ into joint space velocities $\dot{\theta}$ (see Eq.~\ref{eq:joint_space_controller}). $\dot{x}$ is obtained by means of a PID controller (See Eq.~\ref{eq:pos_error} and \ref{eq:task_space_controller}) that attracts the hand position $x$ to the target location $x_{des}$ and a Force Field controller that pushes the hand away from obstacles $x_{obj}$. The controller gains ($K_p, K_i, K_d, K_{rep}$) are heuristically tuned to provide smooth motion and goal convergence. The redundant degrees of freedom are used to maintain a human-like posture while the trajectory is being executed. To do so, a second joint position objective $\theta_{sec}$ is projected into the controller using the nullspace of the Jacobian pseudo-inverse $Null(J^{\dagger})$
\be
	e(t) = x_{des}(t)-x(t)
	\label{eq:pos_error}
\ee
\be
	\dot{x}(t) = K_p e(t)+ K_i \int_0^t e(t) + K_d \frac{de(t)}{dt} + (x(t)-x_{obj}) K_{rep}
	\label{eq:task_space_controller}
\ee
\be
	\dot{\theta}(t) = J^{\dagger}(t) \dot{x}(t) + \text{Null}(J^{\dagger}(t)) (\theta_{sec} - \theta(t))
	\label{eq:joint_space_controller}
\ee

\paragraph{Experiments and Results}
\begin{table}
	\begin{tabularx}{\textwidth}{X|X|X|X}
		{\bf Method } & {\bf Err (cm)} & {\bf Time (ms)}   & {\bf \# Evals} \\
		\hline 
		Grid 1cm   & 1.4 $\pm$ 1.6 & 2.1 $\pm$ 3.2  & 176000\\          
		Grid 10cm  & 6.4 $\pm$ 4.0 & 0.4 $\pm$ 0.4    & 1600\\
        ABC-rej    & 9.1 $\pm$ 11.8 & 912.5 $\pm$ 825.9 & 15000\\
        ABC-SMC    & 10.6$\pm$ 11.2 & 138.9 $\pm$ 275.1 & 15000\\
        MCMC-MH    & 6.5$\pm$ 10.9 & 449.7 $\pm$ 424.2 & 5000\\
		PF         & 2.6 $\pm$ 1.6 & 31.0 $\pm$ 4.7   & 15000\\
	    TP       & 1.5 $\pm$ 1.7 & 16.5 $\pm$ 16.4  & 3937 $\pm$ 6455 \\
	\end{tabularx}
	\caption{Experimental results for accuracy, execution time and \# of likelihood evaluations for different inference strategies. Results obtained using neural likelihood surrogate for 50\% observed trajectory, average after 100 experiments for each inference strategy.}
	\label{tab:reaching}
\end{table}

\begin{figure}
	\centering
	\includegraphics[trim={0cm 1cm 0cm 0cm}, clip, width=0.32\textwidth, height=3.3cm] {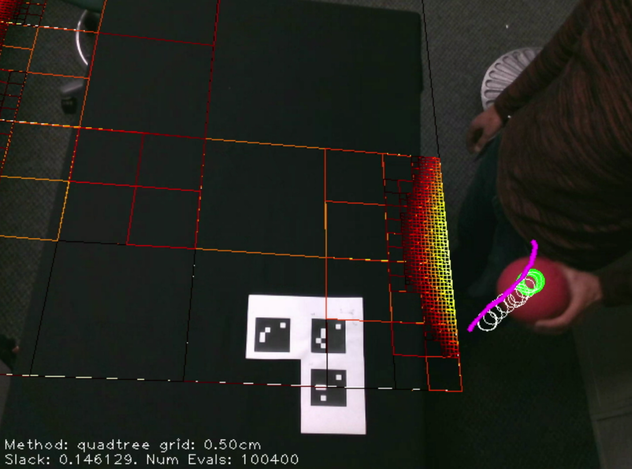}
	\includegraphics[trim={0cm 1cm 0cm 0cm}, clip, width=0.32\textwidth, height=3.3cm] {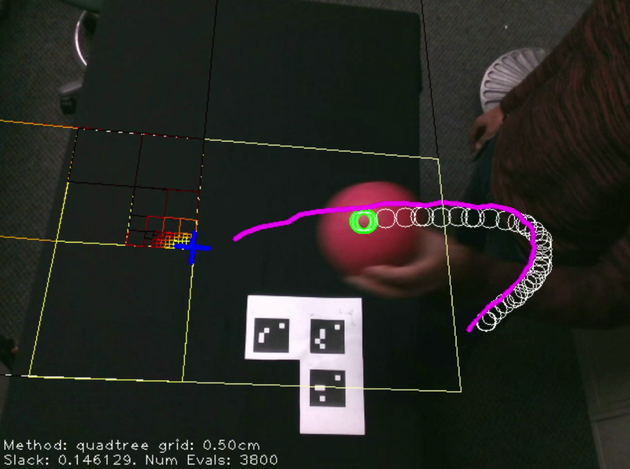}
	\includegraphics[trim={0cm 1cm 0cm 0cm}, clip, width=0.32\textwidth, height=3.3cm] {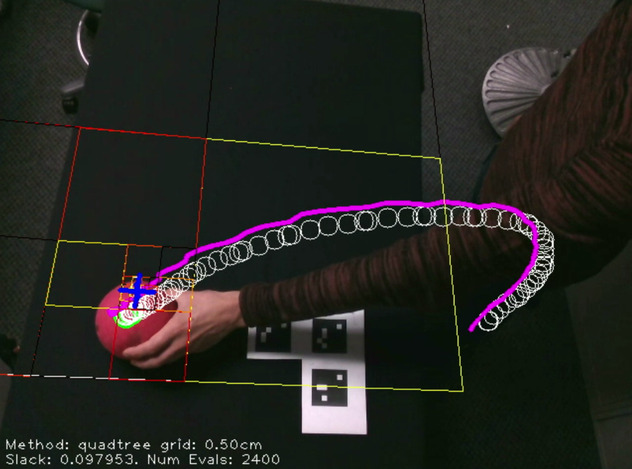}
	\includegraphics[trim={0cm 1cm 0cm 0cm}, clip, width=0.32\textwidth, height=3.3cm]{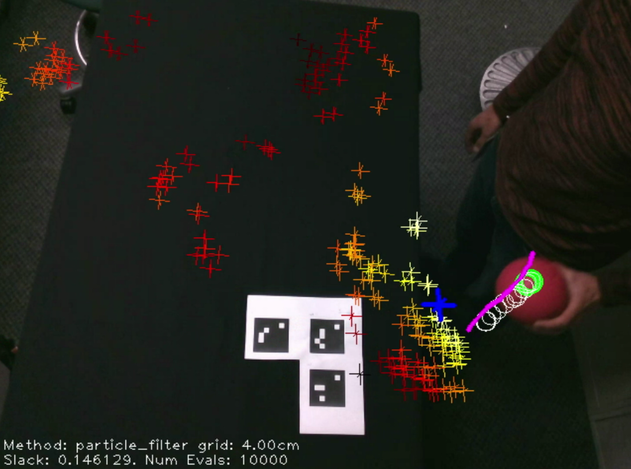}
	\includegraphics[trim={0cm 1cm 0cm 0cm}, clip, width=0.32\textwidth, height=3.3cm]{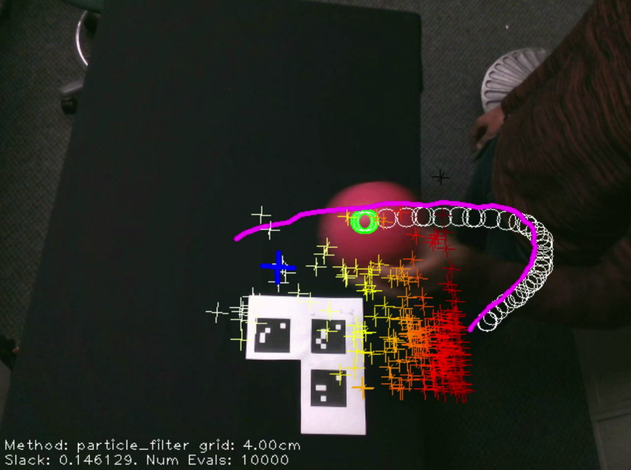}
	\includegraphics[trim={0cm 1cm 0cm 0cm}, clip, width=0.32\textwidth, height=3.3cm]{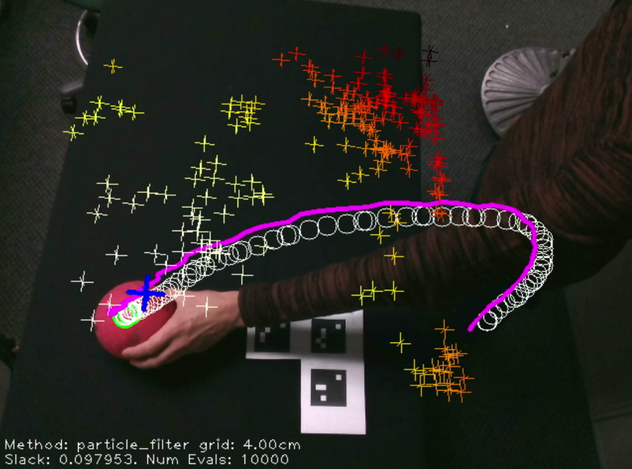}
	\includegraphics[trim={0cm 1cm 0cm 0cm}, clip, width=0.32\textwidth, height=3.3cm]{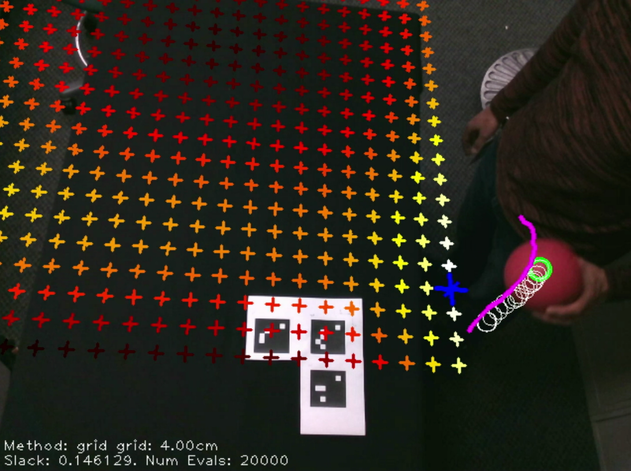}
	\includegraphics[trim={0cm 1cm 0cm 0cm}, clip, width=0.32\textwidth, height=3.3cm]{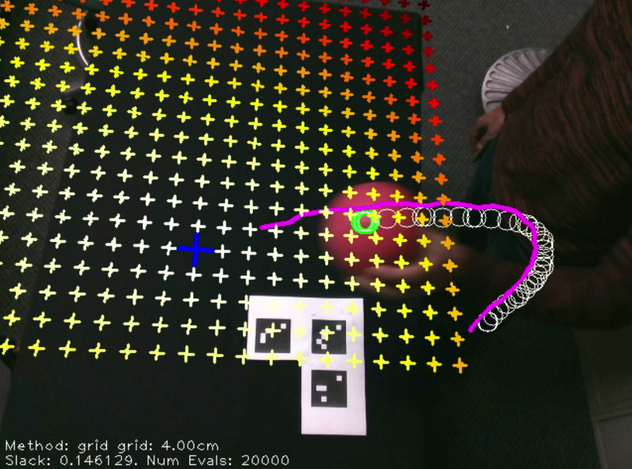}
	\includegraphics[trim={0cm 1cm 0cm 0cm}, clip, width=0.32\textwidth, height=3.3cm]{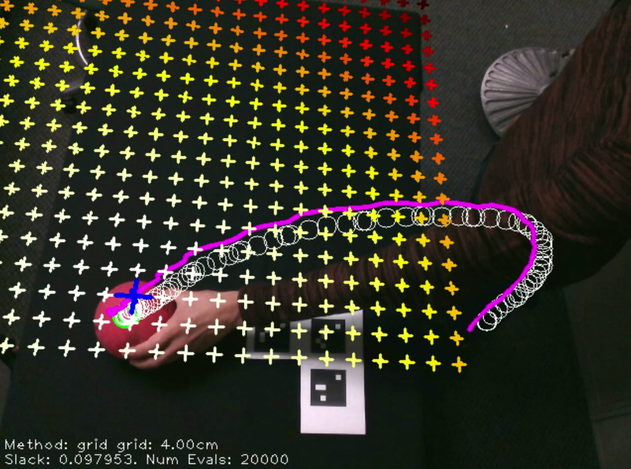}
	\caption{{\bf Qualitative evaluation of inference results for the 2D-TP (top row), particle filter (middle row) and grid (bottom row).} }
	\label{fig:experiments-reaching-qualitative}
\end{figure}

Hand trajectories from real subjects are captured with a Realsense D435 depth camera using a color blob detector and a Kalman filter. These serve as input to the inference procedure. The partially observed $n$ samples of a trajectory are used to compute the posterior by plugging the partially observed trajectory $\tau = D_{0:n-1}$ into Eq.~\ref{eq:surrogate_posterior}: $p(X, \epsilon | \tau) \propto p(X)p(\epsilon)Normal(\tau | f^{*}_{\phi}(X)_{0:n-1}, \epsilon)$.

\begin{figure}
	\centering
	\includegraphics[width=0.95\textwidth]{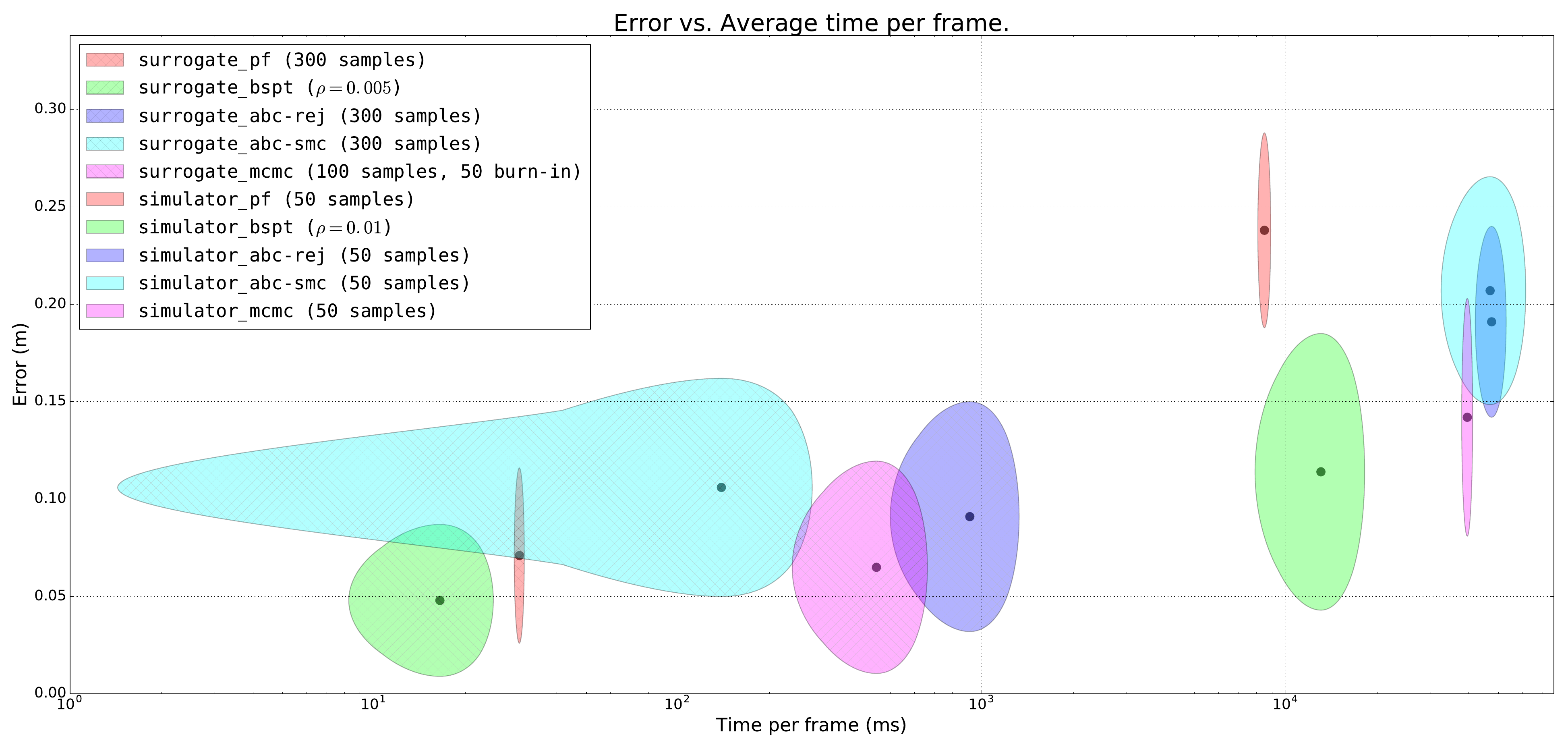}
	\caption{{\bf Error vs. Average time per frame.} Pattern-filled ellipses represent inference methods using neural likelihood surrogate. Ellipses show 1-stdev boundary after averaging 30 experiments. The TP strikes the best accuracy-time.}
	\label{fig:experiments-reaching-quantitative}
\end{figure}

Figure \ref{fig:experiments-reaching-qualitative} shows, qualitatively, the posterior of the inferred goal (with different number of observed trajectory samples) using three types of sampling approaches: a 2D-TP based sampling (ours), particle filter, and a uniform grid based sampling. Figure~\ref{fig:experiments-reaching-quantitative} shows the accuracy vs time results for the proposed inference strategies, including comparison to baseline particle filter~\citep{DelMoral1997}, MCMC Metropolis-Hastings~\citep{Hastings70}, ABC-Reject~\citep{sunnaaker2013approximate} and ABC-SMC~\citep{toni2009approximate}. In our experiments, we use 300 particles to match the computational cost of the adaptive discretization approach. Note that the neural emulator is 30000x-37500x faster than the simulator for this application, thus enabling real-time inference. Note that TP uses $3937 \pm 6455$ likelihood evaluations. This is lower than the $15000$ used by the particle filter baseline, which is also 2x slower than TP. TP also requests fewer evaluations than the $712000$ evaluations used by a 0.5cm grid, which is a grid size that yields competitive accuracy to TP. The TP thus strikes a more appealing runtime/accuracy tradeoff and also scales better in dimensionality than gridding. Detailed quantitative results are shown in Table~\ref{tab:reaching}.

We have performed many simulations, and applied our real-time inference techniques to live data in a real scenario, find an example in the following video: \url{https://bit.ly/2OSNhwA}.

\subsection{Pedestrian crossing behavior model}

The next application on which we illustrate the proposed method is inferring the crossing-intent of pedestrians around a traffic intersection. Without loss of generality, we focus on modeling pedestrians in one corner of the intersection. 

\paragraph{Generative Model}
\begin{figure*}
	\centering
	\includegraphics[width=0.75\textwidth]{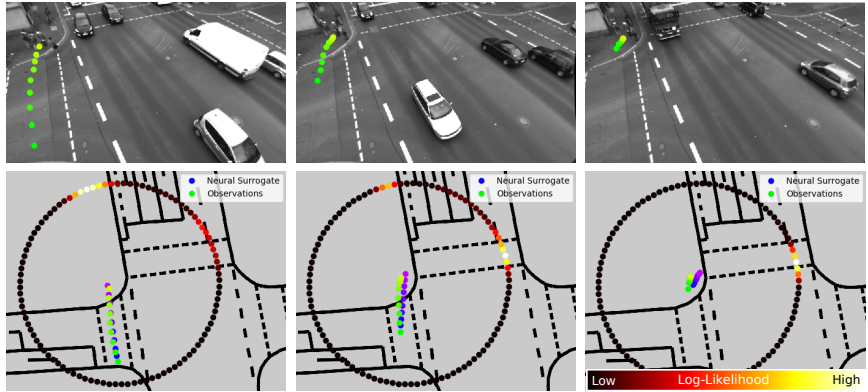}
	\caption{Inferring probable goals of pedestrians in traffic: Each column shows a frame from a real-world scene (top), inferred goal posterior (bottom), observed trajectory (in green/yellow), and a trajectory sampled from the neural surrogate (in blue/purple). In each case, inferred goals qualitatively coincide with common sense.}
	\label{fig:traffic-results}
\end{figure*}

The variable of interest, $X$, is the pedestrian's goal represented as a point on the circumference of a circle representing the current area of interest. Hence it can be represented by an angle in $[0,2\pi]$. The observation $D\in \mathbb{R}^{20}$ is the pedestrian's trajectory ($p_{k+i}\in \mathbb{R}^2, i=\{1,\ldots,10\}$) for 10 consecutive points obtained at 1Hz. Finally, there are a set of nuisance variables	$\eta\in \mathbb{R}^6$ which include: traffic light state $l_k\in\{0,1\} (0-$Red,$1-$Green), time to next traffic light transition $\tau_{k}\in \mathbb{R}$, pedestrian position $p_{k}\in \mathbb{R}^2$, pedestrian speed $v_{k}\in \mathbb{R}$, pedestrian orientation $\theta_k\in [0,2\pi]$. To generate training data we obtained pedestrian trajectories by simulating the traffic in the VISSIM traffic simulator. The neural surrogate, trained with synthetic data from VISSIM, will be used to predict trajectories and the crossing intent of the pedestrians.

\paragraph{Experiments and Results}
Our experiments are applied to a dataset from an intersection in Aschaffenburg, Germany, as collected in~\cite{Strigel2014ko}.  This dataset provides the position (from which speed may be extracted) and orientation of the pedestrian. The traffic light state and the time to the next transition were manually extracted from the video. 

Figure~\ref{fig:traffic-results} shows three frames of the trajectory of a pedestrian approaching an intersection, each 4 seconds apart. In the first frame the pedestrian is crossing the street and approaching a second crossing area; here we obtain the highest likelihood for a pedestrian trajectory continuing straight, with a lower probability for turning right. In the next frame, the trajectory turns slightly right while reducing speed. Here, the highest likelihood is obtained for a trajectory turning right (and waiting for the right of way) and lower for continuing straight. In the last frame, we observe a trajectory turning slightly right and stopping near the crossing area with the highest likelihood for a trajectory crossing through the designated area. 

\begin{figure}
	\def\svgwidth{10cm}
	\begingroup%
	\makeatletter%
	\providecommand\color[2][]{%
		\errmessage{(Inkscape) Color is used for the text in Inkscape, but the package 'color.sty' is not loaded}%
		\renewcommand\color[2][]{}%
	}%
	\providecommand\transparent[1]{%
		\errmessage{(Inkscape) Transparency is used (non-zero) for the text in Inkscape, but the package 'transparent.sty' is not loaded}%
		\renewcommand\transparent[1]{}%
	}%
	\providecommand\rotatebox[2]{#2}%
	\newcommand*\fsize{\dimexpr\f@size pt\relax}%
	\newcommand*\lineheight[1]{\fontsize{\fsize}{#1\fsize}\selectfont}%
	\ifx\svgwidth\undefined%
	\setlength{\unitlength}{460.80003633bp}%
	\ifx\svgscale\undefined%
	\relax%
	\else%
	\setlength{\unitlength}{\unitlength * \real{\svgscale}}%
	\fi%
	\else%
	\setlength{\unitlength}{\svgwidth}%
	\fi%
	\global\let\svgwidth\undefined%
	\global\let\svgscale\undefined%
	\makeatother%
	\begin{picture}(1,1.04145808)%
	\lineheight{1}%
	\setlength\tabcolsep{0pt}%
	\put(0,0){\includegraphics[width=\unitlength,page=1]{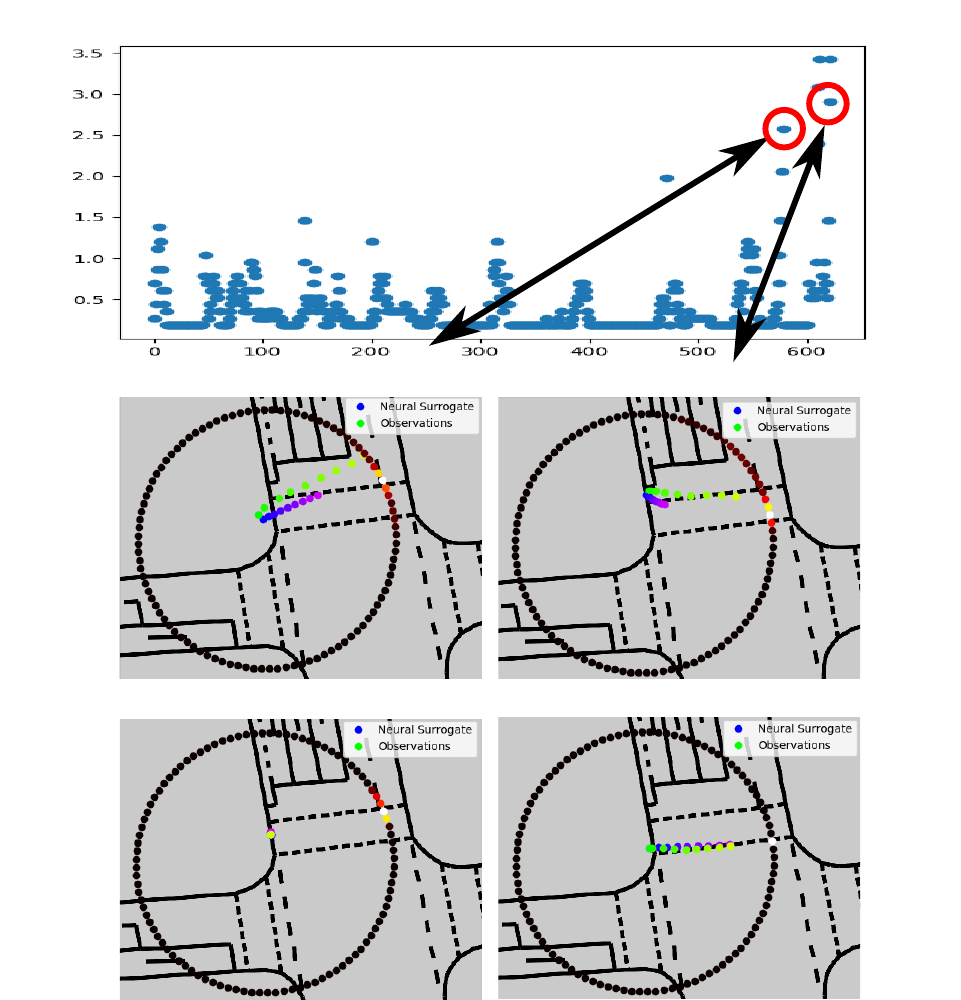}}%
	\scriptsize{
		\put(0.05,0.8){\color[rgb]{0,0,0}\rotatebox{90}{\makebox(0,0)[lt]{\lineheight{1.25}\smash{\begin{tabular}[t]{l}Slack $\epsilon^*$\end{tabular}}}}}%
		\put(0.3,0.64){\color[rgb]{0,0,0}\makebox(0,0)[lt]{\lineheight{1.25}\smash{\begin{tabular}[t]{l}Trajectory Instance from Video\end{tabular}}}}%
		\put(0.115,0.3){\color[rgb]{0,0,0}\makebox(0,0)[lt]{\lineheight{1.25}\smash{\begin{tabular}[t]{l}Trajectory with $\epsilon^*=2.57$\end{tabular}}}}%
		\put(0.115,-0.04){\color[rgb]{0,0,0}\makebox(0,0)[lt]{\lineheight{1.25}\smash{\begin{tabular}[t]{l}Trajectory with $\epsilon^*=0.19$\end{tabular}}}}%
		\put(0.515,0.3){\color[rgb]{0,0,0}\makebox(0,0)[lt]{\lineheight{1.25}\smash{\begin{tabular}[t]{l}Trajectory with $\epsilon^*=2.90$\end{tabular}}}}%
		\put(0.515,-0.04){\color[rgb]{0,0,0}\makebox(0,0)[lt]{\lineheight{1.25}\smash{\begin{tabular}[t]{l}Trajectory with $\epsilon^*=0.29$\end{tabular}}}}%
	}
	\end{picture}%
	\endgroup%
	\centering
	\caption{(Top) Plot showing the inferred slack $\epsilon^*$ (see eq.(\ref{eq:slack})) for different moments in the dataset. (Middle) Two examples of trajectories where $\epsilon^* $ is high, indicating instances where the pedestrian movement deviates from the simulator's predictions. 
	(Bottom) Two examples of trajectories where $\epsilon^*$ is low. 
	}
	\label{Fig:Normal-Abnormal}
\end{figure}
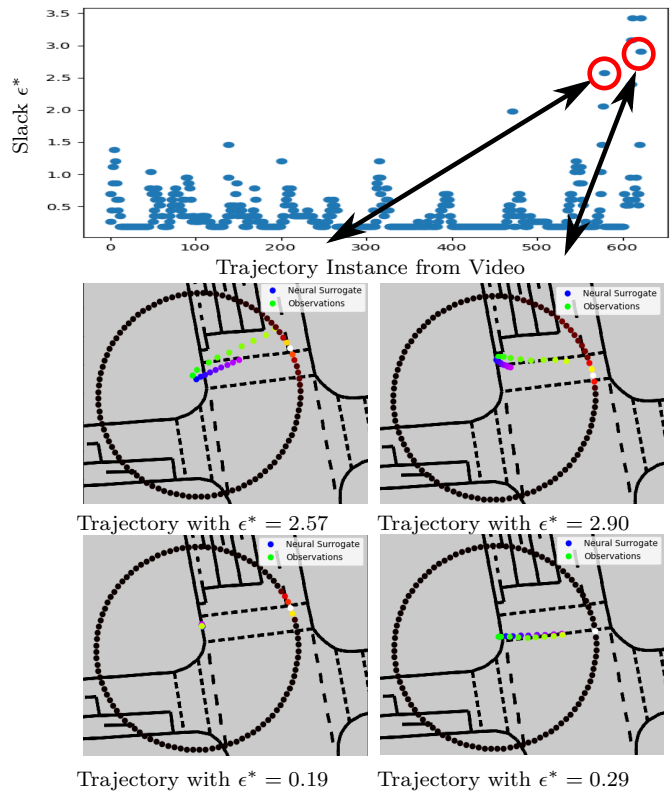

In Figure~\ref{Fig:Normal-Abnormal}, we show two examples to illustrate how $\epsilon$ can be used to detect abnormal behavior of the pedestrians. From the dataset in~\cite{Strigel2014ko} we extracted 1000 instances, each consisting of a 10 seconds observation of a single user. The first example is a person crossing the street with right of way, but outside the designated area. The second one is a pedestrian crossing at the designated area, but without right of way.

\subsection{Pose inference model}
Finally, we demonstrate the application of the proposed method for pose estimation. Here, we  wish to infer the pose of a rigid object from a depth image. Pose is fully specified by 6 degrees of freedom: three for position and three for orientation. Estimating an object's position from a depth image is straightforward. If its pixel coordinates $\mb{p} = (u, v, 1)$ are known, then it's spatial location $\mb{P} = (X_c, Y_c, Z_c)$ relative to the camera can be easily calculated by $\mb{P}=Z_c\mb{K}^{-1}\mb{p}$, where $\mb{K}$ is the camera's intrinsic matrix and $Z_c$ is the depth value from the depth image \citep{hartley2003multiple}. Hence, we focus on estimating orientation only. 

It should be noted that the use of probabilistic techniques to estimate 3D pose is not new; see e.g. ICP (iterative closest point)~\citep{besl1992method} for a classic example. However, the application of real-time ABC is, to the best of our knowledge, novel.

\paragraph{Generative Model}
The generative model consists of the 3D model of the object along with a graphics rendering engine capable of generating depth images from arbitrary viewpoints. The pose is estimated relative to the camera's reference frame (or, equivalently the camera's pose is estimated relative to the object). There are several ways to represent orientation such as Euler angles, rotation matrices, quaternions, etc. We use quaternions as they avoid the problems of ambiguity and gimbal lock associated with Euler angles and lend themselves to efficient representation and computations of rotations. Consider, then, a depth camera capturing the depth image, $D \in \mathbb{R}^{W\times H}$, of an object from an arbitrary viewpoint. $W, H$, are, respectively, the width and height of the image. In a frame of reference attached to the object, the camera's coordinates can be represented in polar form by a set of Euler angles $\Phi = (\phi_1, \phi_2, \phi_3)$. If $\Phi$  follows the Y-X-Z convention, then equivalent quaternion $X=(w,x,y,z)$ can be derived by~\citep{henderson1977euler}:
\begin{equation*}
\begin{aligned}
	w &= \sin \sfrac{\phi_1}{2}\sin \sfrac{\phi_2}{2}\sin \sfrac{\phi_3}{2}+\cos \sfrac{\phi_1}{2}\cos \sfrac{\phi_2}{2}\cos \sfrac{\phi_3}{2}\\
	x &= \sin \sfrac{\phi_1}{2}\cos \sfrac{\phi_2}{2}\sin \sfrac{\phi_3}{2}+\cos \sfrac{\phi_1}{2}\sin \sfrac{\phi_2}{2}\cos \sfrac{\phi_3}{2}\\
	y &= \sin \sfrac{\phi_1}{2}\cos \sfrac{\phi_2}{2}\cos \sfrac{\phi_3}{2}-\cos \sfrac{\phi_1}{2}\sin \sfrac{\phi_2}{2}\sin \sfrac{\phi_3}{2}\\
	z &= \cos \sfrac{\phi_1}{2}\cos \sfrac{\phi_2}{2}\sin \sfrac{\phi_3}{2}-\sin \sfrac{\phi_1}{2}\sin \sfrac{\phi_2}{2}\cos \sfrac{\phi_3}{2}\\
\end{aligned}
\end{equation*}
Although the quaternion is a 4-element vector, it resides on the surface of a 4D unit hypersphere (or, equivalently $X \in SO(3)$ manifold). 

\begin{figure}
	\centering
	\includegraphics[height=2.6cm]{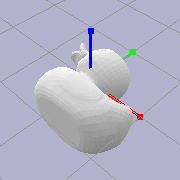}
	\includegraphics[height=2.6cm]{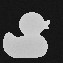}
	\includegraphics[height=2.6cm]{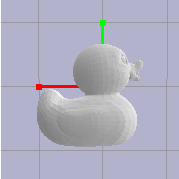}
	\includegraphics[height=2.6cm]{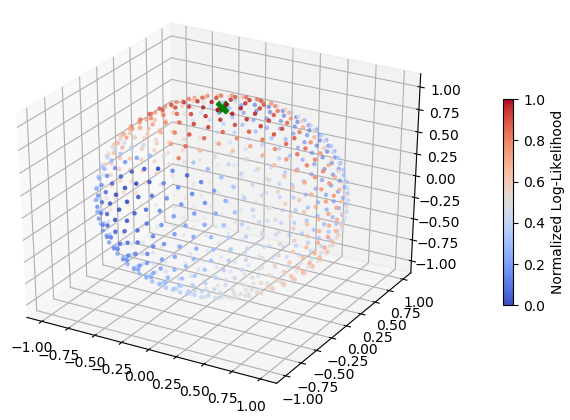}\\
	\includegraphics[height=2.6cm]{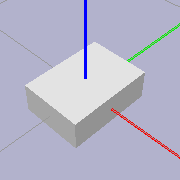}
	\includegraphics[height=2.6cm]{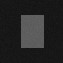}
	\includegraphics[height=2.6cm]{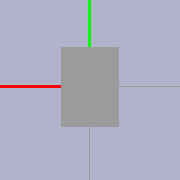} 
	\includegraphics[height=2.6cm]{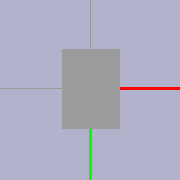}
	\includegraphics[height=2.6cm]{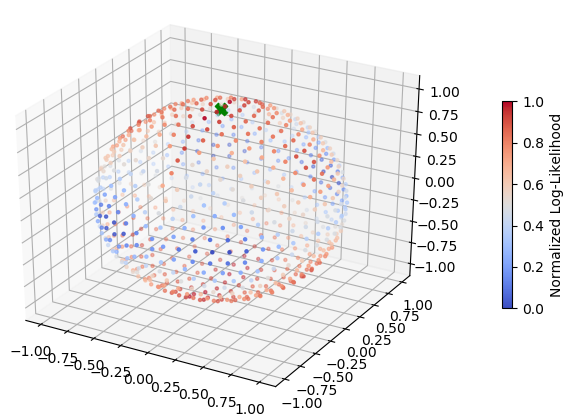}
	
	\caption{Pose inference results for two objects: (Top row) Duck example: 3D Model, sample depth input, view corresponding to a distinct mode for Duck and Log-likelihood plotted over camera viewpoints. (Bottom row) Same for and Box. The green 'X' on the posterior plot is the true camera location.}
	\label{fig:pose-results}
\end{figure}

\paragraph{Experiments and Results}
The proposed approach is tested on two objects: one with symmetries (a rectangular box) and one without (a toy duck). A set of training images is generated overly uniformly and regularly distributed samples over the pose quaternion space. Although the problem of generating evenly-spaced samples on a the surface of a N-D sphere is unsolved in general, approximate computational methods for these exist. Here, we extend one such approach presented in \cite{deserno2004generate} for 3D spheres to 4D spheres. Test inputs are created by generating noisy depth images captured by a virtual depth camera from arbitrary viewpoints. 

Since the pose quaternion is 4-dimensional, a graphical visualizing of its posterior is not possible in general. Hence, for the purposes of display only, results are shown for camera poses that depend on the viewpoint only (i.e, $\phi_3$ is fixed). For both objects, 800 training samples were used. From the results in Figure \ref{fig:pose-results}  it is seen that the pose posterior has two modes for the box object, each of which results in the same observed image. For the duck, whose features are distinctive, the posterior has one mode which matches the observed image.

Finally, we examine the inferred value of slack. For this, we generate 100 images at randomly sampled points over the pose space and note the slack value corresponding to the modes in the posterior. We observe that when the input images are from the same model as the training set, then a low slack value (0.316) is inferred. However, when the input is not from the training distribution, the inferred slack is high (2.511).



	
		

\section{Discussion}
\label{sec:Conclusions}

This paper has shown that it is possible to use ABC for real-time scene understanding. Applications include
inferring the goal to which a person is reaching; inferring the 3D
pose of an object from depth images; and inferring the probable
crossing time of a pedestrian at an intersection. The modeling
approach is based on realistic software simulators that are used as
the core of probabilistic generative models, with a simple Bayesian
error model that links the simulator output to real data. This paper
introduces two techniques for improving performance. The first is the
use of neural surrogates, trained with a simple form of domain
randomization. This paper shows empirically that neural surrogates can
be hundreds to thousands of times faster than the original simulators in our
applications. The second is a technique for adaptive discretization of
the latent variables, that improves inference accuracy and robustness
as compared to Monte Carlo baselines. We hope the techniques
introduced in this paper help researcher explore other real-time
applications of modeling approaches from the Approximate Bayesian
Computation community, and to develop new real-time methods for approximate Bayesian inference.

%



\bibliographystyle{unsrt}
\bibliography{bibliography}

\end{document}